%% file: main-arxiv-preprint.tex
\documentclass{article} 
\usepackage{preprint,times}
\iclrfinalcopy

\input{math_commands.tex}

\usepackage{url}

\definecolor{iccvblue}{rgb}{0.21,0.49,0.74}
\usepackage[colorlinks,linkcolor=red,citecolor=iccvblue]{hyperref}
\usepackage{multirow}
\usepackage{colortbl}
\usepackage{graphicx}
\usepackage{booktabs}
\usepackage{xcolor}
\usepackage{verbatim}
\usepackage{arydshln}
\usepackage{array}
\usepackage{makecell}
\usepackage{caption}
\usepackage{wrapfig}
\usepackage{subcaption}
\usepackage{amssymb}
\usepackage{pifont}
\newcommand{\tabincell}[2]{\begin{tabular}{@{}#1@{}}#2\end{tabular}}

\title{LD-RSVIS: A Large-Scale and Diverse \\Benchmark for Referring Surgical Video \\Instrument Segmentation}

\author{\textbf{Zan Wang$^{1}$}\quad
\textbf{Yunhe Feng$^{1}$}\quad
\textbf{Dong Nie$^{2}$}\quad
\textbf{Oluwatosin Oluwadare$^{1}$}\quad
\textbf{Kewei Sha$^{1}$}\\
\textbf{Yan Huang$^{1}$}\quad
\textbf{Heng Fan$^{1}$}\\
$^{1}$University of North Texas, Denton, USA \quad
$^{2}$Meta Inc, USA
}

\begin{document}

\maketitle

\begin{abstract}

Referring surgical video instrument segmentation (RSVIS) aims at segmenting the instrument in
a surgical video, given a textual description. Despite recent progress, current models are trained and assessed on relatively small-scale benchmarks, hindering the development of more general RSVIS. In addition, existing benchmarks support only the single-target expression that refers to one instrument in the video, while overlooking multi-target and no-target referring expressions, restricting the applicability of RSVIS in practical scenarios. Addressing these issues, we propose \textbf{\emph{LD-RSVIS}}, a new benchmark aiming to facilitate more robust and general RSVIS. Specifically, LD-RSVIS consists of 3,536 surgical videos with 1.09 million frames and covers a broad set of 30 instrument classes from 25 various procedures. By including abundant videos and classes, LD-RSVIS could benefit large-scale training and evaluation of more general RSVIS methods. Besides, unlike existing datasets, LD-RSVIS offers diverse referring settings, including no-target, single-target, and multi-target expressions, which enables the development of more practical RSVIS models in real applications. In order to ensure high-quality annotations, all videos in LD-RSVIS are manually labeled with multiple rounds of inspection and refinement. To our knowledge, LD-RSVIS is the largest and most diverse benchmark for RSVIS. To analyze LD-RSVIS and to provide comparison for future research, we evaluate 12 representative methods, and the results  reveal that more efforts are required for improvements. To encourage future research, we present a simple yet effective RSVIS method, dubbed \textbf{\emph{Cascade-RSVIS}}, that first mines target-specific cues using the complementary multi-cue text information and then employs such cues and textual information for segmentation, achieving promising performance. Through LD-RSVIS and Cascade-RSVIS, we hope to inspire more future research toward robust and general RSVIS. Our benchmark and code will be released \href{https://github.com/zanwang01/LD-RSVIS}{here}.
\end{abstract}

\section{Introduction}

\setlength{\columnsep}{8pt}%
\setlength\intextsep{0pt}
\begin{wrapfigure}{r}{0.483\textwidth}
    \centering
    \includegraphics[width=1\linewidth]{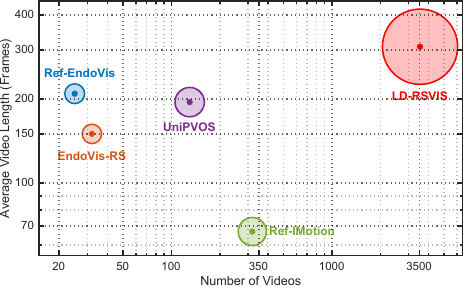}\vspace{-0mm}
    \caption{Comparison of existing RSVIS benchmarks. LD-RSVIS is larger than all other benchmarks. \emph{Best viewed in color for all figures}.}
    \label{fig:comp_scale}
\end{wrapfigure}

Referring surgical video instrument segmentation (RSVIS) aims at segmenting an instrument of interest within a surgical sequence, given the textual expression. Owing to its important applications in various scenarios, including surgical robotics and intelligent surgical assistants, RSVIS has received increasing interest in recent years~\citep{zhou2023text,wang2024video,liu2025resurgsam2,wei2026moves}. In particular, several datasets have been introduced~\citep{wang2024video,liu2025resurgsam2,wei2026moves,liu2026unisurgsam} for training and assessing RSVIS models, greatly facilitating RSVIS. Despite this, these benchmarks suffer from several critical limitations that hinder further development of RSVIS.

One of the major issues with existing RSVIS benchmarks is their relatively small scale. Particularly, in the deep learning era, abundant surgical sequences are crucial for learning robust RSVIS models. However, as displayed in Fig.~\ref{fig:comp_scale}, all current RSVIS datasets contain \emph{fewer than} 350 video sequences, making them \emph{inadequate} for large-scale model training and thereby hindering the further advances in RSVIS. Beyond facilitating model training, a large-scale benchmark is also essential for reliable and comprehensive evaluation. Nevertheless, existing RSVIS datasets comprise \emph{less than} 100 sequences in their test sets, making the evaluation susceptible to bias and potentially sensitive to a small number of individual videos. Moreover, such limited test sets in current benchmarks may fail to adequately capture diverse scenarios in the real-world surgeries. Consequently, the reported  performance may not faithfully reflect the true capability and generality of RSVIS methods.

\begin{figure*}[t]
    \centering
    \includegraphics[width=0.97\textwidth]{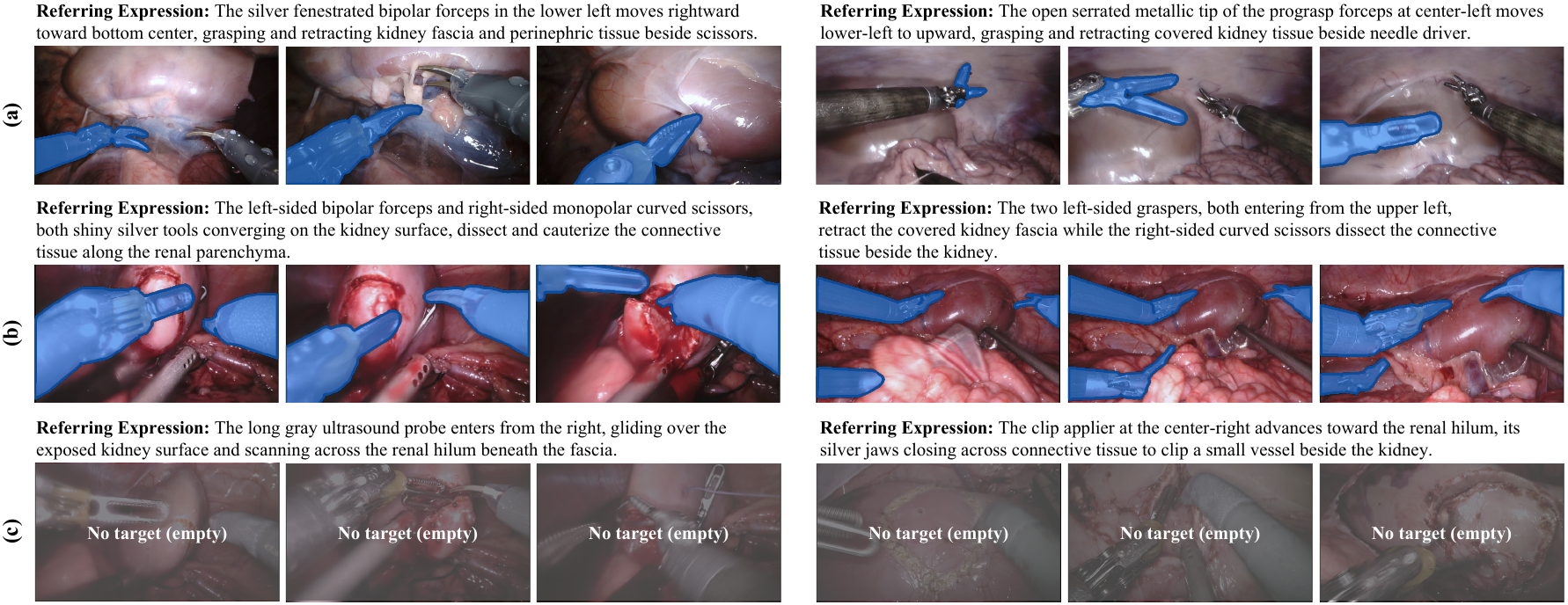}\vspace{-1mm}
    \caption{Illustration of different referring settings in RSVIS. Image (a): Single-target expression in LD-RSVIS and other benchmarks; Image (b): Multi-target expression in our LD-RSVIS; Image (c): No-target expression in LD-RSVIS. All examples are from LD-RSVIS.}\vspace{-3mm}
    \label{fig:referring-settings}
\end{figure*}

Besides the small-scale problem, another limitation of current benchmarks is the restricted referring setting. Specifically, existing benchmarks only support the single-target expression (see Fig.~\ref{fig:referring-settings} (a)), which restricts RSVIS models to predicting one and only one target. In practical scenarios, however, a referring expression may simultaneously describe multiple instruments (see Fig.~\ref{fig:referring-settings} (b)), particularly when they are involved in the same surgical interaction, requiring the model to identify and segment all matching targets. Furthermore, in certain situations, the target of interest may not appear in the video sequence at all (see Fig.~\ref{fig:referring-settings} (c)), requiring the model to recognize the absence of the referred target, instead of incorrectly matching the expression to an irrelevant target. To better reflect diverse referring needs in real-world surgeries, a benchmark should support more comprehensive referring settings, including no-target, single-target, and multi-target expressions.

\textbf{Contribution.} To alleviate the above limitations and to further facilitate RSVIS, we present a Large-scale and Diverse benchmark for RSVIS, named \textbf{\emph{LD-RSVIS}}. Specifically, LD-RSVIS contains 3,536 surgical video sequences with 1.09 million frames, and covers a broad set of 30 instrument categories from 25 various procedures. Compared to current benchmarks, our LD-RSVIS is 10$\times$ larger in scale, enabling more robust training and more faithful evaluation of RSVIS approaches. In addition, unlike existing datasets that support only the single-target expressions, LD-RSVIS, inspired by generalized referring image segmentation~\citep{liu2023gres}, provides more diverse referring settings, including no-target expressions, where the target of interest is absent; single-target expressions, where the text refers to a single target; and multi-target expressions, where text refers to multiple target objects. In total, LD-RSVIS provides 58,666 textual expressions across these three referring settings, aiming at facilitating the development of more practical RSVIS systems. To ensure the high annotation quality, all videos in LD-RSVIS are manually labeled and undergo multiple rounds of careful inspection and refinement. To our best knowledge, LD-RSVIS is currently the \emph{largest} RSVIS benchmark, and also the \emph{first} to introduce various referring  settings into RSVIS.

In order to understand how existing method perform and to provide comparisons for future research, we extensively evaluate 12 representative methods that provide executable implementations, including recent RSVIS approaches and general referring video object segmentation (RVOS) models, and conduct in-depth comparison and analysis. Our evaluation results (as described later) display that, not surprisingly, these top-performing approaches perform substantially below satisfactory levels on the more challenging LD-RSVIS benchmark. This demonstrates the difficulties in achieving robust and generic referring surgical video instrument segmentation in real-world scenarios, and more efforts are required to improve the performance for practical applications.

Moreover, to facilitate future research, we further introduce a simple yet effective method for RSVIS, dubbed \textbf{\emph{Cascade-RSVIS}}. Specifically, considering that the given expression alone is \emph{insufficient} to describe the targets in a video involving severe appearance variations or distractors, Cascade-RSVIS, built on SAM 2~\citep{ravi2025sam}, employs a two-stage strategy, consisting of target-context mining (TCM) and context-guided segmentation (CGS), for RSVIS. More concretely, TCM aims to generate target-specific appearance cues in a video using textual information encoded in referring expression, while CGS leverages such target appearance cues to enhance the memory and further incorporates the textual feature for more accurate target segmentation. By connecting TCM and CGS in a unified framework, Cascade-RSVIS is able to benefit from additional discriminative target appearance cues, enabling more robust segmentation. In addition, to better exploit target-relevant information from the expression, we present a multi-cue text representation that combines the global textual semantics with learned complementary semantic cues, and applies it in both TCM and CGS, further improving the performance. Despite simplicity, Cascade-RSVIS achieves promising results on LD-RSVIS, and is expected to provide a solid reference for future research.

In summary, our main contributions are as follows: \ding{171} We introduce LD-RSVIS, a large-scale dataset comprising 3,536 surgical videos with 1.09 million frames and diverse referring settings for RSVIS; \ding{170} We evaluate 12 representative methods to understand their performance and to offer comparisons to future research; \ding{168} We propose a simple yet effective method, Cascade-RSVIS, to facilitate more future research on RSVIS; \ding{169} Cascade-RSVIS shows promising
results on LD-RSVIS, establishing a solid reference and providing guidance for future research.

\section{Related Work}
\label{sec:related_work}

\noindent\textbf{RSVIS Benchmarks.} Benchmarks play an important role for the development of RSVIS. EndoVis-RS~\citep{wang2024video} is the first benchmark for referring surgical video instrument segmentation. It is constructed from the popular EndoVis challenges~\citep{Allan2019EndoVis17,Allan2020EndoVis18} and consists of 32 sequences with 4.8K frames. Ref-EndoVis~\citep{liu2025resurgsam2} is built upon EndoVis-RS~\citep{wang2024video} with refined annotations by addressing inconsistencies and omissions in the original masks, comprising 25 videos and 5.2K frames. Later in~\citep{liu2026unisurgsam}, Ref-EndoVis is further extended into a unified dataset UniPVOS by including extra video sequences. Notably, both Ref-EndoVis and UniPVOS extend the scope of referring surgical video segmentation beyond surgical instruments by additionally incorporating tissues as referred targets. More recently, Ref-IMotion~\citep{wei2026moves} offers 319 video sequences with 21K frames to facilitate motion-aware RSVIS. Despite these efforts, existing datasets remain limited in scale, hindering further advances in RSVIS. To alleviate this, LD-RSVIS provides 3,536 videos with 1.09M frames, which is 10$\times$ larger than current datasets. In addition, unlike existing benchmarks supporting only single-target expressions, LD-RSVIS offer more diverse referring settings, including no-target, single-target, and multi-target expressions, for more practical RSVIS. Tab.~\ref{tab:RSVIS-benchmarks} summarizes the differences between LD-RSVIS and existing datasets.

\renewcommand{\arraystretch}{1.05}
\begin{table*}[!t]
\setlength{\tabcolsep}{3.5pt}
    \centering
    \caption{Comparison of our LD-RSVIS and current RSVIS datasets. ``NT'', ``ST'', and ``MT'' indicate no-target, single-target, and multi-target, respectively. ``n/a'' denotes that data is not available. The video frame rate of all datasets is 1 frame per second (fps).}\vspace{-3mm}
    \resizebox{0.99\textwidth}{!}{
        \begin{tabular}{rc*{6}{c}*{3}{>{\centering\arraybackslash}p{0.9cm}}}
        \Xhline{1.2pt}
        & & & & & & & & \multicolumn{3}{c}{\textbf{Referring Expression}} \\
        \cmidrule(lr){9-11}
        \multirow{-2}{*}{\textbf{Benchmark}} & \multirow{-2}{*}{\textbf{Year}} & \multirow{-2}{*}{\textbf{Videos}} & \multirow{-2}{*}{\tabincell{c}{\textbf{Total}\\\textbf{Frames}}} & \multirow{-2}{*}{\tabincell{c}{\textbf{Instru.}\\\textbf{Classes}}} & \multirow{-2}{*}{\tabincell{c}{\textbf{Avg. Vid.}\\\textbf{Length}}} & \multirow{-2}{*}{\tabincell{c}{\textbf{Num. of}\\\textbf{Expressions}}} & \multirow{-2}{*}{\tabincell{c}{\textbf{Num. of}\\\textbf{Masks}}} & \textbf{ST} & \textbf{MT} & \textbf{NT} \\
        \hline\hline
        EndoVis-RS~\citep{wang2024video} & 2024 & 32 & 4.8\textbf{K} & 9 & 150 & n/a & 9.8\textbf{K} & \ding{51} & \ding{55} & \ding{55} \\
        Ref-EndoVis~\citep{liu2025resurgsam2} & 2025 & 25 & 5.2\textbf{K} & 9 & 209 & n/a & 12.3\textbf{K} & \ding{51} & \ding{55} & \ding{55} \\
        UniPVOS~\citep{liu2026unisurgsam} & 2026 & 130 & 25.0\textbf{K} & n/a & 195 & 1,367 & 178.8\textbf{K} & \ding{51} & \ding{55} & \ding{55} \\
        Surg-IMotion~\citep{wei2026moves} & 2026 & 319 & 21.0\textbf{K} & n/a & 67 & 718 & n/a & \ding{51} & \ding{55} & \ding{55} \\
        \hline
        \rowcolor[HTML]{e9f7ef} \textbf{LD-RSVIS} (Ours) & 2026 & 3,536 & 1.09\textbf{M} & 30 & 308 & 58,666 & 3.85\textbf{M} & \ding{51} & \ding{51} & \ding{51} \\
        \Xhline{1.2pt}
        \end{tabular}}
    \label{tab:RSVIS-benchmarks}\vspace{-3mm}
\end{table*}

\textbf{RSVIS Algorithms.} RSVIS has recently drawn growing interest owing to its important applications in surgical robotics and intelligent surgical assistants. The method of~\citep{zhou2023text} introduces a text-promptable architecture for surgical instrument segmentation by leveraging text-based category prompts and multimodal vision-language models. The work of~\citep{wang2024video} proposes a video-instrument synergistic learning method for text-referred instrument segmentation. The work of~\citep{liu2025resurgsam2} introduces the credible initial-frame selection and diversity-driven memory to improve SAM 2~\citep{ravi2025sam} for RSVIS, which is further extended in~\citep{liu2026unisurgsam} by unifying multiple prompts. The method of~\citep{wei2026moves} presents a motion-aware framework for RSVIS. Different from these methods, Cascade-RSVIS introduces a two-stage architecture that first mines target-specific cues from the video and then employs such cues for segmentation, enabling the exploration of additional discriminative target appearance cues for robust performance.

\textbf{Referring Video Object Segmentation.} Referring video object segmentation (RVOS) aims to segment the target given a natural language description, and has been extensively studied over the past decade with many datasets~\citep{gavrilyuk2018actor,khoreva2018video,seo2020urvos,ding2023mevis,liang2026long} and methods~\citep{wu2022language,wu2023onlinerefer,luo2023soc,liang2025referdino,pan2025semantic,Cheng_2026_CVPR,Jiang_2026_CVPR}. Different from these works, we specifically focus on segmenting referred instruments in surgical videos, where frequent occlusions, visually similar distractors, and repeated target entry and exit pose substantial difficulties for accurate target identification and segmentation. To advance research in this direction, we present a new benchmark for referring surgical instrument segmentation, together with an effective baseline method.

\section{The Proposed LD-RSVIS Benchmark}

\subsection{Construction Principle}
LD-RSVIS aims to offer a dedicated dataset to foster the development of RSVIS. For this purpose, we follow several principles in constructing LD-RSVIS: 
\textbf{(i) \emph{Large-scale}.} Large-scale benchmark is crucial for training robust RSVIS models and for providing comprehensive evaluation. Therefore, we expect LD-RSVIS to contain at least 3,000 videos, benefiting both large-scale model training and assessment. \textbf{(ii) \emph{Diverse referring settings.}} In real-world surgical scenarios, the referring expression may describe one or several instruments or have no matching target. To this end, LD-RSVIS aims to offer more diverse referring settings, including no-target, single-target, and multi-target expression, enable more practical RSVIS. \textbf{(iii) \emph{Rich procedures and instrument categories}.} An important goal of LD-RSVIS is to advance the development of general RSVIS which aims at segmenting instruments of different classes in various procedures. To this end, the new benchmark is designed to cover a broad range of instrument categories and surgical procedures. \textbf{(iv) \emph{High quality.}} High-quality annotations are essential for both training and evaluation. To ensure the high quality of LD-RSVIS, all sequences are manually labeled and undergo multiple rounds of careful inspection and refinement.

\subsection{Data Acquisition}

Similar to existing RSVIS benchmarks~\citep{wang2024video,liu2025resurgsam2,liu2026unisurgsam,wei2026moves}, we construct LD-RSVIS by collecting publicly available videos from current surgical video datasets. However, unlike prior RSVIS datasets that primarily gather videos from surgical video segmentation datasets, we broaden the source pool to contain a wider range of surgical video understanding tasks, thereby increasing the diversity of surgical instruments and procedures in our dataset. Specifically, we source videos from datasets developed for surgical scene segmentation and/or detection~\citep{Allan2019EndoVis17,Allan2020EndoVis18,murali2023endoscapes,ayobi2025pixel,wang2022autolaparo,maier2021heidelberg,yoon2021hsdb}, surgical phase and action recognition~\citep{rueckert2026video,lavanchy2024challenges,bawa2021saras,psychogyios2023sar}, surgical workflow understanding~\citep{twinanda2016endonet,wagner2023comparative}, surgical task and activity recognition~\citep{zia2026surgical}, and large-scale surgical video pre-training~\citep{JASPERS2026103873}. Together, these diverse sources offer a broad basis for constructing a more general large-scale and diverse RSVIS benchmark. Due to space limitation, we present the data selected from each source in the \emph{\textbf{appendix}}.

Once the source pool is determined, we conduct a verification process in which the videos from each source dataset are carefully inspected to assess their suitability for the RSVIS task. This process is carried out by several experts (students who work on related surgical video understanding tasks) to ensure the quality of selected videos. Specifically, for each sequence, we first remove the corrupted or irrelevant content. If the remaining video comprises at least one clip suitable for RSVIS, we retain this sequence; otherwise, it is discarded. Since many of the original surgical videos are considerably long, often spanning tens of minutes to several hours, we follow a similar strategy to existing RSVIS benchmarks and sample several representative clips rather than using the entire video. These selected clips are chosen to capture diverse instrument appearances and classes, surgical stages, and temporal conditions, reducing temporal redundancy while preserving diversity of the original surgical videos. Due to space limitation, we show statistics of selected clips from each video source in the \emph{\textbf{appendix}}.

Eventually, we compile a large-scale benchmark, LD-RSVIS, for RSVIS. LD-RSVIS contains 3,536 videos with 1.09 million frames and covers 30 instrument classes from 25 procedures. Each video is paired with multiple referring expressions covering different referring scenarios, including no-target, single-target, and multi-target cases, resulting in a total of 58,666 expressions. Compared to current RSVIS datasets, LD-RSVIS is much larger and more diverse. Tab.~\ref{tab:RSVIS-benchmarks} summarizes LD-RSVIS and existing benchmarks. Due to limited space, additional details on the instrument classes and surgical procedures, as well as discussions of dataset maintenance, ethical considerations, and responsible usage of the proposed LD-RSVIS, are provided in the \emph{\textbf{appendix}}.

\subsection{Annotation}

After data collection, our experts first generate referring expressions for each sequence. The referring expressions can describe instruments at different levels of granularity, covering both complete instruments and their semantic parts, enabling more fine-grained and flexible referring segmentation. Different from current RSVIS benchmarks, we, inspired by generalized referring image segmentation~\citep{liu2023gres}, provide different types of expressions, including no-target, single-target, and multi-target referring expressions. In a no-target expression, the referred target is absent; in a single-target expression, the text refers to a single target; in a multi-target expression, the text refers to a varying number (larger than one) of target objects. It is worth noting that, for videos containing only one valid target, multi-target expressions are not provided. In total, 58,666 referring expressions are generated in LD-RSVIS. Afterwards, we label the corresponding targets referred in the expressions from videos with consistent spatio-temporal masklets.

\begin{figure*}[!t]
    \centering
    \begin{minipage}[b]{0.245\linewidth}
        \centering
        \includegraphics[width=\linewidth]{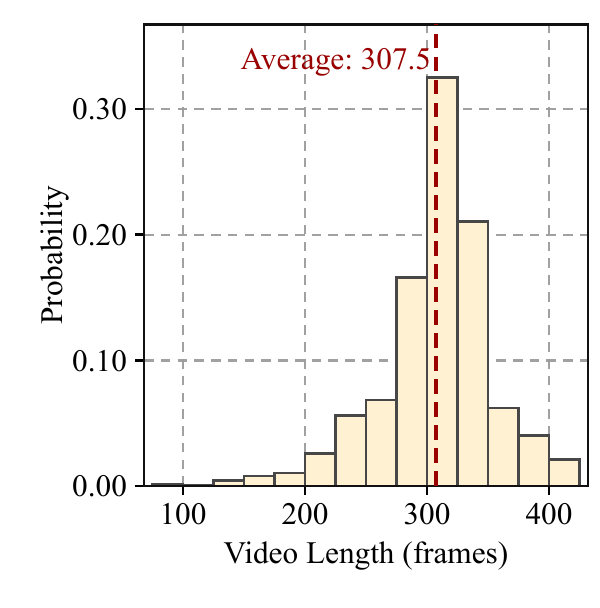}\vspace{-2.5mm}
        \captionsetup{labelformat=empty, font={scriptsize}, skip=1pt}
        \subcaption*{(a) Distribution of video length}
        \label{subfig:sequence-length-statistics}
    \end{minipage}\hfill
    \begin{minipage}[b]{0.245\linewidth}
        \centering
        \includegraphics[width=\linewidth]{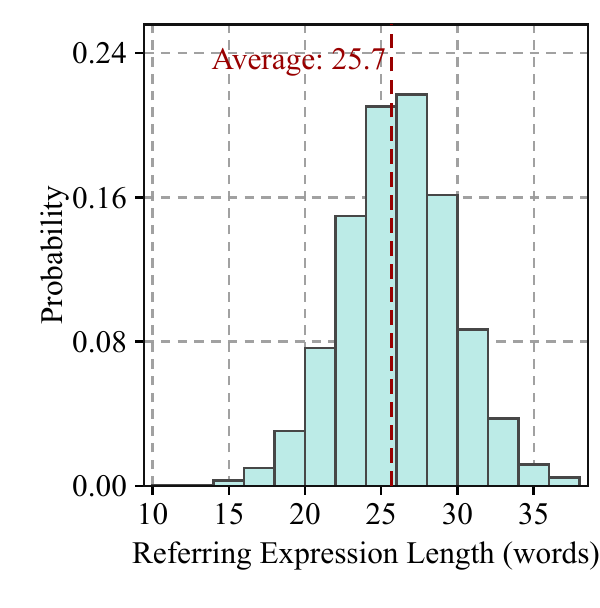}\vspace{-2.5mm}
        \captionsetup{labelformat=empty, font={scriptsize}, skip=1pt}
        \subcaption*{(b) Distribution of expression length}
        \label{subfig:query-length-statistics}
    \end{minipage}\hfill
    \begin{minipage}[b]{0.245\linewidth}
        \centering
        \includegraphics[width=\linewidth]{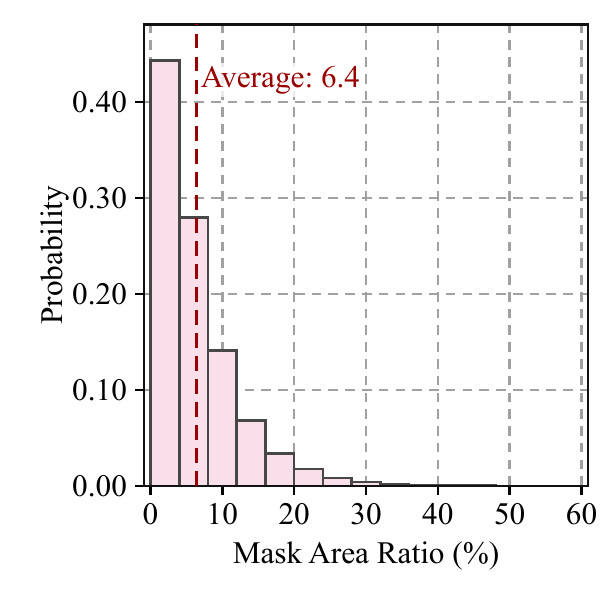}\vspace{-2.5mm}
        \captionsetup{labelformat=empty, font={scriptsize}, skip=1pt}
        \subcaption*{(c) Distribution of mask area ratio}
        \label{subfig:mask-area-statistics}
    \end{minipage}\hfill
    \begin{minipage}[b]{0.245\linewidth}
        \centering
        \includegraphics[width=\linewidth]{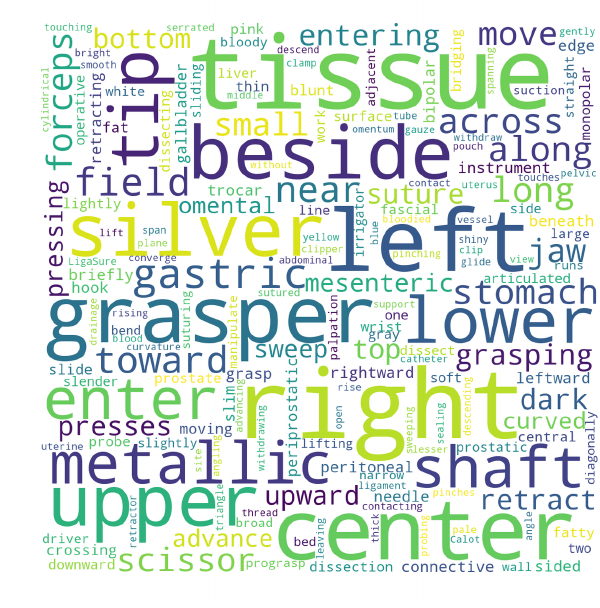}\vspace{-2.5mm}
        \captionsetup{labelformat=empty, font={scriptsize}, skip=1pt}
        \subcaption*{(d) Word cloud of referring expressions}
        \label{subfig:query-wordcloud}
    \end{minipage}
    \vspace{-2mm}
    \caption{Representative statistics of LD-RSVIS, comprising distributions of video length, referring expression length, mask area ratio, and wordcloud of referring expressions.}
    \label{fig:annotation-statistics}
    \vspace{-4mm}
\end{figure*}

To ensure high-quality annotations of LD-RSVIS, we compile an annotation team, composed of an expert group (students working on related surgical video understanding tasks) for verification and a volunteer group for labeling, and use a multi-round annotation strategy. In specific, the expert group first label the referred targets in initial frames. Then, the volunteer group, divided into several teams, works to annotate the complete masklet based on the initial mask for each referring expression. After this
initial round, the masklet annotations are sent to the expert group for inspection, where the mask completeness and boundaries, temporal consistency, and target identity will be carefully reviewed. If the initial annotations are not unanimously agreed by all experts, they will be returned back to the original labeling team for refinement. We repeat this
process until all annotations satisfy the quality requirements. Due to space limitation, we show the pipeline
of annotation process in the \emph{\textbf{appendix}}. Fig.~\ref{fig:referring-settings} shows several annotation examples in LD-RSVIS, and more can be seen in the \emph{\textbf{appendix}}.

\textbf{Statistics of annotation.} To better understand LD-RSVIS, we demonstrate representative statistics in Fig.~\ref{fig:annotation-statistics}, including the distributions of the video length, referring expression length, mask area ratio (defined as the percentage of pixels occupied by referred targets in the corresponding frame), and wordcloud of referring expressions. As displayed in Fig.~\ref{fig:annotation-statistics}, LD-RSVIS contains videos and referring expressions of varying lengths and target objects of different scales. Together, these statistics exhibit the diversity of LD-RSVIS, making it a challenging and realistic benchmark for RSVIS.

\subsection{Dataset Split and Evaluation Metric}

\setlength{\columnsep}{2pt}%
\setlength\intextsep{0pt}
\begin{wraptable}{r}{0.5\textwidth}
\setlength{\tabcolsep}{3pt}
	\centering
	\renewcommand{\arraystretch}{1.1}
        \caption{Comparison of training and test sets.}\vspace{-2mm}
  \resizebox{0.46\textwidth}{!}{
    \begin{tabular}{cccccc}
    \Xhline{1.2pt}
    & \textbf{Videos} & \tabincell{c}{\textbf{Total}\\ \textbf{Frames}} & \tabincell{c}{\textbf{Avg.}\\ \textbf{Length}} & \tabincell{c}{\textbf{Num. of}\\ \textbf{Expre.}} &\tabincell{c}{\textbf{Num. of}\\ \textbf{Masks}} \\
    \hline\hline
    LD-RSVIS$_\text{Tra}$   & 2,853 & 880,080 & 308 & 47,500 & 3,135,866 \\
    LD-RSVIS$_\text{Tst}$   & 683 & 207,282 & 303 & 11,166 & 711,398 \\
    \Xhline{1.2pt}
    \end{tabular}}
    \label{tab:splitset}
 \vspace{5pt}
\end{wraptable}

\textbf{Dataset Split.} LD-RSVIS contains 3,536 videos, with 2,853  sequences selected for the training set, named LD-RSVIS$_{\text{Tra}}$, and the rest 683 for test set, dubbed LD-RSVIS$_{\text{Tst}}$. To prevent train-test leakage, video clips derived from the same original long surgical video are assigned exclusively to either training set or test set. Besides, both LD-RSVIS$_{\text{Tra}}$ and LD-RSVIS$_{\text{Tst}}$ follow the same referring settings, comprising no/single/multi-target expressions. Tab.~\ref{tab:splitset} summarizes training and test sets.

\noindent\textbf{Evaluation Metrics.} Following~\citep{liu2025resurgsam2}, we apply two metrics for evaluation, including region similarity $\mathcal{J}$, which computes
the Intersection over Union (IoU) of the predicted and ground-truth masks, and contour accuracy $\mathcal{F}$~\citep{perazzi2016benchmark}, which evaluates the contour accuracy of prediction results. For comprehensive evaluation, we further calculate the average of $\mathcal{J}$ and $\mathcal{F}$, denoted as $\mathcal{J}\&\mathcal{F}$. It is worth noting that, following~\citep{DingLHYJLJ25}, for a no-target expression, a prediction is considered correct only if the output remains empty throughout the entire sequence. In this case, both $\mathcal{J}$ and $\mathcal{F}$ are set to 1; otherwise, both are set to 0. Due to limited space, please refer to~\citep{perazzi2016benchmark,DingLHYJLJ25,liu2025resurgsam2} for more details of these metrics.

\section{The Proposed Cascade-RSVIS}
\label{sec:cascade-rsvis}

\begin{figure*}[t]
    \centering
    \includegraphics[width=0.95\textwidth]{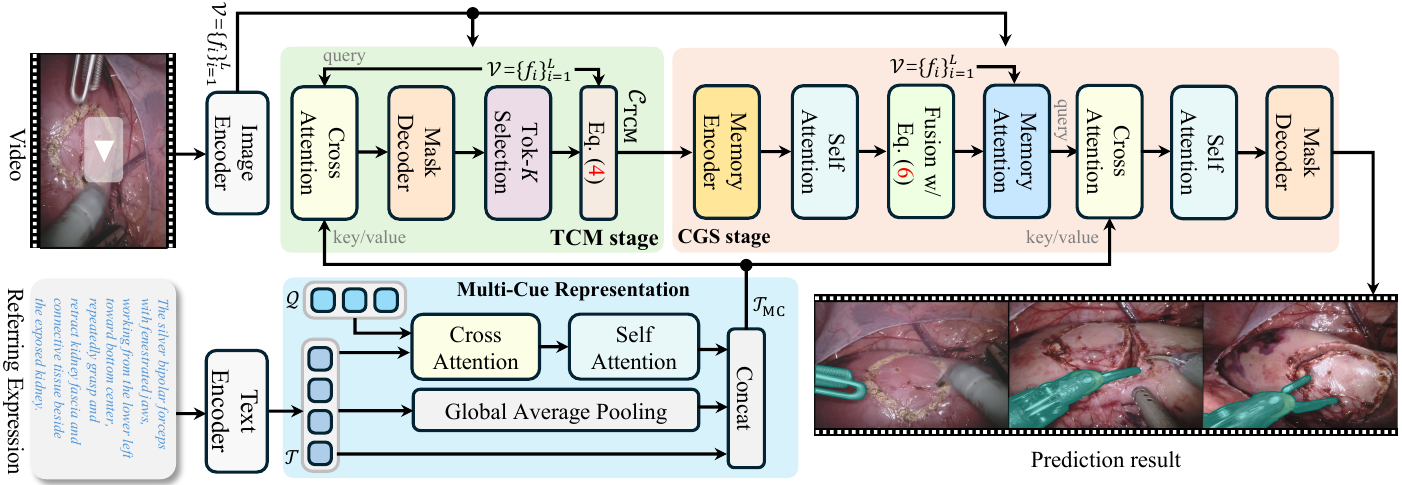}\vspace{-1.5mm}
    \caption{Overview of Cascade-RSVIS that first mines target-relevant context via TCM and then leverages such context together with the referring expression for segmentation through CGS.}
    \label{fig:cascade-rsvis-overview}\vspace{-2mm}
\end{figure*}

\textbf{Overview.}
We propose Cascade-RSVIS for RSVIS by extending SAM~2~\citep{ravi2025sam}, which exploits target-specific appearance cues from a video through a two-stage framework. As in Fig.~\ref{fig:cascade-rsvis-overview}, after feature extraction, the target-context mining (TCM) stage first mines reliable target context using the referring expression, and the context-guided segmentation (CGS) stage then leverages such context together with the referring expression for segmentation.

\subsection{Feature Extraction}
\label{sec:feature-extraction}

In Cascade-RSVIS, we first extract visual and textual features for the video and referring expression.

\textbf{Video Feature Extraction.} Given a video with $L$ frames $\mathcal{I}=\{I_i\}_{i=1}^{L}$, we apply the image encoder $\mathrm{\Phi}(\cdot)$ from SAM~2~\citep{ravi2025sam} to extract visual features $\mathcal{V}$ via $\mathcal{V}=\{f_i\}_{i=1}^{L}$, where $f_i=\mathrm{\Phi}(I_i)$. The visual features will be used in the subsequent stages for target context mining and segmentation.

\textbf{Multi-Cue Textual Feature Extraction for Expression.} Referring expressions often contain multiple complementary cues to identify the target, which may not be fully captured by a single textual representation. Hence, we construct a multi-cue textual feature that combines word-level, holistic, and learned complementary semantics. Specifically, as in Fig.~\ref{fig:cascade-rsvis-overview}, we first tokenize the expression into $\mathcal{W}=\{w_j\}_{j=1}^{N_t}$ and employ  RoBERTa~\citep{liu2019roberta} to obtain word embeddings $\mathcal{T}=\{t_j\}_{j=1}^{N_t}$. We then perform average pooling over $\mathcal{T}$ to obtain a holistic sentence feature $\tilde{\mathcal{T}}=\mathtt{AvgPool}(\mathcal{T})$, which summarizes overall semantics of the  expression. To further capture fine-grained complementary cues in the expression, we introduce $N_q$ learnable queries $\mathcal{Q}=\{q_n\}_{n=1}^{N_q}$. These queries retrieve information from the word sequence through cross-attention and subsequently interact via self-attention, yielding $\mathcal{U}=\mathtt{SA}(\mathtt{CA}(\mathcal{Q},\mathcal{T}))$, where $\mathtt{SA}(\textbf{z})$ denotes the self-attention block with $\textbf{z}$ generating query/key/value, and $\mathtt{CA}(\textbf{z}, \textbf{u})$ the cross-attention block with $\textbf{z}$ generating query and $\textbf{u}$ key/value, as in~\citep{VaswaniSPUJGKP17}. Finally, we concatenate the word feature $\mathcal{T}$, sentence feature $\tilde{\mathcal{T}}$, and query features $\mathcal{U}$ to produce the final multi-cue textual representation $\mathcal{T}_{\mathrm{MC}}=\mathtt{Concat}(\mathcal{T},\tilde{\mathcal{T}},\mathcal{U})$.

\subsection{Our Cascade-RSVIS Framework}
\label{sec:cascade-framework}

After extracting video and expression features, Cascade-RSVIS performs segmentation through two cascaded stages, including TCM and CGS, as illustrated in the following.

\textbf{Target-Context Mining (TCM).} The TCM stage mines target-relevant context from the video based on the referring expression. To this end, we first inject the textual feature into each frame through cross-attention followed by a residual connection to obtain the multimodal video features $\mathcal{V}_{\text{TCM}}$ via
\begin{equation}
\setlength{\abovedisplayskip}{5pt} 
\setlength{\belowdisplayskip}{5pt}
\mathcal{V}_{\text{TCM}}=\{f_i^{\text{TCM}}\}_{i=1}^{L} \;\;\;\;\;\; f_i^{\text{TCM}}=\mathtt{CA}(f_i,\mathcal{T}_{\mathrm{MC}})+f_i
\end{equation}
where $\mathtt{CA}(\cdot, \cdot)$ is the cross-attention block. Afterwards, we apply mask decoder $\mathtt{MaskDec(\cdot)}$ in~\citep{ravi2025sam} to each $f_i^{\text{TCM}}$ of $\mathcal{V}_{\text{TCM}}$, as follows,
\begin{equation}
\setlength{\abovedisplayskip}{5pt} 
\setlength{\belowdisplayskip}{5pt}
    M_i^{\text{TCM}}=\mathtt{MaskDec}(f_i^{\text{TCM}})
\end{equation}
where $M_i^{\text{TCM}}$ denotes the prediction in frame $i$, and contains a mask $m_{i}^{\text{TCM}}$, an IoU score $s_{i}^{\text{TCM}}$, and an object-presence score $p_{i}^{\text{TCM}}$. 

Given the predicted results $R^{\text{TCM}}=\{M_i^{\text{TCM}}\}_{i=1}^{L}$ over the video, we select target context from these predictions. Following SAM~2~\citep{ravi2025sam}, if the maximum object-presence score of $R^{\text{TCM}}$ is below zero, we directly return an empty mask sequence. Otherwise, we retain predictions with non-negative object-presence scores and select the top $K$ indices according to their predicted IoU scores, as follows,
\begin{equation}
\setlength{\abovedisplayskip}{5pt} 
\setlength{\belowdisplayskip}{5pt}
    \mathcal{K}=\mathtt{IdxTopK}_{i\in \mathcal{D}}(s_{i}^{\text{TCM}}) \;\;\;\;\;\; \mathcal{D}=\{i | p_{i}^{\text{TCM}} \ge 0\}
\end{equation}
where $\mathcal{K}$ denotes indices of the selected top $K$ predictions, and $\mathtt{IdxTopK}(\cdot)$ returns indices of the top $K$ predictions ranked by IoU scores. Based on $\mathcal{K}$, the target context $\mathcal{C}_{\text{TCM}}$ is formed as follows,
\begin{equation}
\setlength{\abovedisplayskip}{5pt} 
\setlength{\belowdisplayskip}{5pt}
    \mathcal{C}_{\mathrm{TCM}}=\{ (f_i,m_i^{\text{TCM}},s_i^{\text{TCM}})\}_{i\in \mathcal{K}}
\end{equation}
The resulting $\mathcal{C}_{\mathrm{TCM}}$ is then passed to the subsequent CGS stage for segmentation.

\textbf{Context-Guided Segmentation (CGS).} The CGS stage aims to leverage the context $\mathcal{C}_{\mathrm{TCM}}$ from TCM to guide segmentation. Specifically, for each selected context in $\mathcal{C}_{\mathrm{TCM}}$, we first generate its memory by encoding its video feature and mask with memory encoder $\mathtt{MemEnc}(\cdot,\cdot)$ in SAM~2 via
\begin{equation}
\setlength{\abovedisplayskip}{5pt} 
\setlength{\belowdisplayskip}{5pt}
    \mathcal{M}_i = \mathtt{MemEnc}(f_i,m_i^{\text{TCM}}) \;\;\; i \in \mathcal{K}
\end{equation}
Once obtaining $\mathcal{M}_{\text{CGS}}=\{\mathcal{M}_i\}_{i \in \mathcal{K}}$, we perform self-attention on it for feature interaction via $\bar{\mathcal{M}}_{\text{CGS}}=\{\bar{\mathcal{M}}_i\}_{i\in \mathcal{K}}=\mathtt{SA}(\mathcal{M}_{\text{CGS}})$ with $\bar{\mathcal{M}_i}$ the memory feature in frame $i$ after interaction, and then aggregate memory features in $\Bar{\mathcal{M}}_{\text{CGS}}$ based on their normalized IoU scores, as follows,
\begin{equation}
\setlength{\abovedisplayskip}{5pt} 
\setlength{\belowdisplayskip}{5pt}
    \tilde{\mathcal{M}}_{\text{CGS}} = \sum\nolimits_{i \in \mathcal{K}}\omega_i \cdot \bar{\mathcal{M}_i} \;\;\;\;\;\;\;\; \omega_i = \mathtt{exp}(s_i^{\text{TCM}})/\sum\nolimits_{j\in\mathcal{K}}\mathtt{exp}(s_j^{\text{TCM}})
\end{equation}
where $\tilde{\mathcal{M}}_{\text{CGS}}$ denotes the fused memory feature derived from the context $\mathcal{C}_{\text{TCM}}$, which is injected into the video features via memory attention $\mathtt{MemAtt}(\cdot, \cdot)$ in SAM 2, as follows,
\begin{equation}
\setlength{\abovedisplayskip}{5pt} 
\setlength{\belowdisplayskip}{5pt}
    \mathcal{V}_{\text{CGS}}=\{f_i^{\text{CGS}}\}_{i=1}^{L} \;\;\;\;\;\; f_i^{\text{CGS}}=\mathtt{MemAtt}(f_i,\tilde{\mathcal{M}}_{\text{CGS}})
\end{equation}
With memory-enhanced video features $\mathcal{V}_{\text{CGS}}$, we further inject the expression feature into each $f_i^{\text{CGS}}$ to obtain the multimodal feature $g_i^{\text{CGS}}$ through cross-attention followed by a residual connection via 
\begin{equation}
\setlength{\abovedisplayskip}{5pt} 
\setlength{\belowdisplayskip}{5pt}
g_i^{\text{CGS}}=\mathtt{CA}(f_i^{\text{CGS}},\mathcal{T}_{\text{MC}})+f_i^{\text{CGS}}
\end{equation}
With $\mathcal{G}=\{g_i^{\text{CGS}}\}_{i=1}^{L}$, we further perform self-attention to model temporal interaction in the video via $\tilde{\mathcal{G}}=\{\tilde{g}_i^{\text{CGS}}\}_{i=1}^{L}=\mathtt{SA}(\mathcal{G})$, where $\mathtt{SA(\cdot)}$ represents the self-attention block. Afterwards, we apply the mask decoder $\mathtt{MaskDec}(\cdot)$ to each $\tilde{g}_i^{\text{CGS}}$ and retain its predicted mask, as follows,
\begin{equation}
\setlength{\abovedisplayskip}{5pt} 
\setlength{\belowdisplayskip}{5pt}
m_i^{\text{CGS}} = \mathtt{MaskDec}(\tilde{g}_i^{\text{CGS}})
\end{equation}
where $m_i^{\text{CGS}}$ denotes predicted mask segmentation in each frame $i$.

\textbf{Optimization.} In Cascade-RSVIS, the segmentation predictions from both TCM and CGS are optimized during training. Due to space limitations, the loss functions are provided in the \textbf{\emph{appendix}}.

\begin{table*}[!t]
    \setlength{\tabcolsep}{6pt}
    \centering
    \renewcommand{\arraystretch}{1.075}
    \caption{Comparison with existing methods on LD-RSVIS$_{\text{Tst}}$. ``MTE'', ``STE'', and ``NTE'' indicate evaluations on multi-target, single-target, and no-target referring expressions, respectively.}\vspace{-2mm}
    \resizebox{0.98\textwidth}{!}{
        \begin{tabular}{rcccccccccc}
    \specialrule{1.5pt}{0pt}{0pt}
     & \multicolumn{3}{c}{\textbf{Overall}} & \multicolumn{3}{c}{\textbf{MTE}} & \multicolumn{3}{c}{\textbf{STE}} & \textbf{NTE} \\
    \cmidrule(lr){2-4} \cmidrule(lr){5-7} \cmidrule(lr){8-10} \cmidrule(lr){11-11}
          & $\mathcal{J}$     & $\mathcal{F}$     & $\mathcal{J}\&\mathcal{F}$  & $\mathcal{J}$    & $\mathcal{F}$     & $\mathcal{J}$\&$\mathcal{F}$  & $\mathcal{J}$     & $\mathcal{F}$     & $\mathcal{J}$\&$\mathcal{F}$  & $\mathcal{J}$\&$\mathcal{F}$ \\
    \hline\hline
    VIS-Net~\citep{wang2024video}\textsubscript{[T-MI'24]} & 20.8  & 18.6  & 19.7  & 19.7  & 15.5  & 17.6  & 24.4  & 22.1  & 23.3  & 14.7 \\
    ReferFormer~\citep{wu2022language}\textsubscript{[CVPR'22]} & 19.0  & 23.6  & 21.3  & 23.3  & 31.9  & 27.6  & 21.2  & 26.1  & 23.7  & 10.3 \\
    OnlineRefer~\citep{wu2023onlinerefer}\textsubscript{[ICCV'23]} & 20.4  & 24.2  & 22.3  & 27.8  & 34.0  & 30.9  & 21.5  & 26.0  & 23.7  & 10.9 \\
    LoSh~\citep{yuan2024losh}\textsubscript{[CVPR'24]}  & 22.5  & 28.1  & 25.3  & 26.4  & 38.8  & 32.6  & 26.0  & 31.0  & 28.5  & 11.7 \\
    LMPM~\citep{ding2023mevis}\textsubscript{[ICCV'23]}  & 25.5  & 27.3  & 26.4  & 37.4  & 39.0  & 38.2  & 21.0  & 23.8  & 22.4  & 22.7 \\
    ReSurgSAM2~\citep{liu2025resurgsam2}\textsubscript{[MICCAI'25]} & 27.0  & 29.6  & 28.3  & 29.1  & 33.1  & 31.1  & 26.4  & 29.6  & 28.0  & 26.1 \\
    SurgRef~\citep{wei2026moves}\textsubscript{[AAAI'26]} & 29.4  & 27.8  & 28.6  & 38.2  & 36.0  & 37.1  & 27.8  & 25.7  & 26.8  & 23.8 \\
    DsHmp~\citep{he2024decoupling}\textsubscript{[CVPR'24]} & 24.7  & 33.3  & 29.0  & 28.1  & 46.9  & 37.5  & 22.9  & 30.7  & 26.8  & 25.0 \\
    SAMWISE~\citep{cuttano2025samwise}\textsubscript{[CVPR'25]} & 28.8  & 32.2  & 30.5  & 32.2  & 38.0  & 35.1  & 33.8  & 37.7  & 35.8  & 15.4 \\
    ReferDINO~\citep{liang2025referdino}\textsubscript{[ICCV'25]} & 29.2  & 33.6  & 31.4  & 30.1  & 37.5  & 33.8  & 34.7  & 39.8  & 37.3  & 17.3 \\
    ReferMo~\citep{liang2026long}\textsubscript{[CVPR'26]} & 31.6  & 32.0  & 31.8  & 40.6  & 36.4  & 38.5  & 36.7  & 39.6  & 38.2  & 12.4 \\
    FlowRVS~\citep{wang2026deforming}\textsubscript{[ICLR'26]} & 29.8  & 36.4  & 33.1  & 31.1  & 46.3  & 38.7  & 36.0  & 41.6  & 38.8  & 16.2 \\
    \hline
    \rowcolor[HTML]{e9f7ef}
    Cascade-RSVIS (ours) & 36.1  & 38.7  & 37.4  & 37.8  & 41.8  & 39.8  & 38.0  & 41.2  & 39.6  & 30.6 \\
    \specialrule{1.5pt}{0pt}{0pt}
    \end{tabular}%
        }
    \label{tab:mainres}
    \vspace{-2mm}
\end{table*}

\section{Experiments}

\textbf{Implementation of Cascade-RSVIS.} Cascade-RSVIS is implemented in PyTorch~\citep{paszke2019pytorch} and trained on two NVIDIA A100 (40G) GPUs. The image encoder $\Phi(\cdot)$ adopts the Hiera-B+~\citep{ryali2023hiera}. Cascade-RSVIS is trained in two stages: TCM for 18 epochs and CGS for 24 epochs, with a batch size of 2 using AdamW~\citep{LoshchilovH19}. We first optimize TCM and then freeze it while training CGS. The initial learning rates are set to $5\times10^{-6}$ for pretrained parameters and $1\times10^{-4}$ for newly introduced modules. During inference, the clip length $L$ is set to 8 due to memory constraints, and predictions from all clips are combined to obtain the final video segmentation results. The parameters $N_q$ and $K$ are both empirically set to 3.

\textbf{Evaluated Methods.} We evaluate 12 representative approaches with publicly available implementations on LD-RSVIS. These contain three surgical RSVIS methods, including VIS-Net~\citep{wang2024video}, ReSurgSAM2~\citep{liu2025resurgsam2}, and SurgRef~\citep{wei2026moves}, and nine general RVOS models, including ReferFormer~\citep{wu2022language}, OnlineRefer~\citep{wu2023onlinerefer}, LMPM~\citep{ding2023mevis}, LoSh~\citep{yuan2024losh}, DsHmp~\citep{he2024decoupling}, SAMWISE~\citep{cuttano2025samwise}, ReferDINO~\citep{liang2025referdino}, ReferMo~\citep{liang2026long}, and FlowRVS~\citep{wang2026deforming}. It is worth noting that, all the evaluated methods are trained on LD-RSVIS$_{\text{Tra}}$ for fair comparison.

\subsection{Evaluation Results}

\textbf{Overall evaluation.} We demonstrate results of evaluated approaches and Cascade-RSVIS in
Tab.~\ref{tab:mainres}. Our Cascade-RSVIS achieves the best overall $\mathcal{J}$, $\mathcal{F}$, and $\mathcal{J}\&\mathcal{F}$ scores of 36.1\%, 38.7\%, and 37.4\%, respectively. Specifically, in terms of $\mathcal{J}\&\mathcal{F}$, it outperforms the second best method, FlowRVS with 33.1\% $\mathcal{J}\&\mathcal{F}$ score, by 4.3\%. Cascade-RSVIS also improves the overall $\mathcal{J}$ score of ReferMo by 4.5\%  and the overall $\mathcal{F}$ score of FlowRVS by 2.3\%. These results consistently validate the efficacy of our Cascade-RSVIS by mining and leveraging target context from videos for segmentation.

\textbf{Evaluation across different expressions.} To enable in-depth analysis of methods, we show results across multi-target (2,732 expressions), single-target (6,573 expressions), and no-target referring expressions (1,861 expressions) in Tab.~\ref{tab:mainres}. Please note that, for no-target expressions, $\mathcal{J}$, $\mathcal{F}$, and $\mathcal{J}$\&$\mathcal{F}$ are identical because both $\mathcal{J}$ and $\mathcal{F}$ are set to 1 for an entirely empty prediction and 0 otherwise; therefore, we report only $\mathcal{J}$\&$\mathcal{F}$. As shown in Tab.~\ref{tab:mainres}, Cascade-RSVIS shows the best $\mathcal{J}$\&$\mathcal{F}$ scores of 39.8\%, 41.2\%, and 30.6\% on multi-, single-, and no-target expressions, showing its effectiveness.

\begin{table*}[!t]
\centering
\setlength{\tabcolsep}{6.5pt}
\begin{minipage}[t]{0.49\textwidth}
\vspace{0pt}
\centering
\renewcommand{\arraystretch}{1.05}
\caption{Ablation on the cascade framework.}
\vspace{-2mm}
\resizebox{\linewidth}{!}{
\begin{tabular}{rcccc}
            \specialrule{1.5pt}{0pt}{0pt}
            & Framework & $\mathcal{J}$ & $\mathcal{F}$ & $\mathcal{J}\&\mathcal{F}$ \\
            \hline\hline
            \ding{182} & TCM alone & 28.9 & 32.5 & 30.7 \\
            \rowcolor{cyan!10}
            \ding{183} & Cascade TCM and CGS & 36.1 & 38.7 & 37.4 \\
            \specialrule{1.5pt}{0pt}{0pt}
        \end{tabular}}
    \label{tab:stage-ablation}
\end{minipage}
\hfill
\begin{minipage}[t]{0.49\textwidth}
\vspace{0pt}
\centering
\renewcommand{\arraystretch}{1.05}
\setlength{\tabcolsep}{9.5pt}
\caption{Ablation on $K$ in TCM.}
\vspace{-2mm}
\resizebox{\linewidth}{!}{
\begin{tabular}{rcccc}
            \specialrule{1.5pt}{0pt}{0pt}
            & Top $K$ predictions & $\mathcal{J}$ & $\mathcal{F}$ & $\mathcal{J}\&\mathcal{F}$ \\
            \hline\hline
            \ding{182} & $K=1$ & 34.9 & 37.5 & 36.2 \\
            \rowcolor{cyan!10}
            \ding{183} & $K=3$ & 36.1 & 38.7 & 37.4 \\
            \ding{184} & $K=5$ & 35.9 & 38.3 & 37.1 \\
            \specialrule{1.5pt}{0pt}{0pt}
        \end{tabular}}
    \label{tab:topk-ablation}
\end{minipage}
\hfill
\begin{minipage}[t]{0.49\textwidth}
\vspace{-1pt}
\centering
\renewcommand{\arraystretch}{1.05}
\caption{Ablation on multi-cue representation.}
\vspace{-2mm}
\resizebox{\linewidth}{!}{
\begin{tabular}{rcccccc}
            \specialrule{1.5pt}{0pt}{0pt}
            & \tabincell{c}{Word \\ feat.} & \tabincell{c}{Sentence \\ feat.} & \tabincell{c}{Query \\ feat.} & $\mathcal{J}$ & $\mathcal{F}$ & $\mathcal{J}\&\mathcal{F}$ \\
            \hline\hline
            \ding{182} & \checkmark & - & - & 33.7 & 36.3 & 35.0 \\
            \ding{183} & \checkmark & \checkmark & - & 34.6 & 37.6 & 36.1 \\
            \ding{184} & \checkmark & - & \checkmark & 35.2 & 38.0 & 36.6 \\
            \rowcolor{cyan!10}
            \ding{185} & \checkmark & \checkmark & \checkmark & 36.1 & 38.7 & 37.4 \\
            \specialrule{1.5pt}{0pt}{0pt}
        \end{tabular}}
    \label{tab:text-ablation}
\end{minipage}
\hfill
\begin{minipage}[t]{0.49\textwidth}
\vspace{5pt}
\centering
\renewcommand{\arraystretch}{1.03}
\setlength{\tabcolsep}{6.5pt}
\caption{Ablation on context fusion in CGS.}
\vspace{-2mm}
\resizebox{\linewidth}{!}{
\begin{tabular}{rcccc}
            \specialrule{1.5pt}{0pt}{0pt}
            & \tabincell{c}{Context \\ fusion strategy} & $\mathcal{J}$ & $\mathcal{F}$ & $\mathcal{J}\&\mathcal{F}$ \\
            \hline\hline
            \ding{182} & Addition & 34.8 & 37.2 & 36.0 \\
            \ding{183} & Gated fusion & 35.6 & 38.2 & 36.9 \\
            \rowcolor{cyan!10}
            \ding{184} & IoU-weighted addition & 36.1 & 38.7 & 37.4 \\
            \specialrule{1.5pt}{0pt}{0pt}
        \end{tabular}}
    \label{tab:fusion-ablation}
\end{minipage}\vspace{-2mm}
\end{table*}

\subsection{Ablation Study on Cascade-RSVIS}

We conduct ablations for Cascade-RSVIS on LD-RSVIS$_{\text{Tst}}$. Our final setting is highlighted in {\color{cyan!75} cyan}.

\textbf{Ablation on cascade framework.} 
Cascade-RSVIS connects TCM and CGS stages in a cascaded manner, where TCM mines target context and CGS exploits it for segmentation. As TCM can directly produce frame-level masks, we compare its standalone predictions with the complete cascade in Tab.~\ref{tab:stage-ablation}. As in Tab.~\ref{tab:stage-ablation}, TCM alone achieves 30.7\% $\mathcal{J}$\&$\mathcal{F}$ score (\ding{182}). When cascading TCM and CGS, the $\mathcal{J}$\&$\mathcal{F}$ score is significantly improved to 37.4\% with by 6.7\% (\ding{183}) owing to the exploration of context as a guidance for segmentation. This substantial improvement validates the effectiveness of the proposed cascaded framework.

\textbf{Ablation on $K$ in TCM.} We select the top $K$ predictions in TCM to obtain the target context $\mathcal{C}_{\text{TCM}}$. We conduct an ablation on $K$ in Tab.~\ref{tab:topk-ablation}. As shown in Tab.~\ref{tab:topk-ablation}, $K=3$ achieves the best $\mathcal{J}$\&$\mathcal{F}$ of 37.4\% (\ding{183}), while a larger $K=5$ slightly degrades the performance (\ding{184}), likely due to less reliable target cues and background noise. We therefore set $K$ to 3.

\textbf{Ablation on multi-cue  representation for expression.} We construct a multi-cue representation to extract expression feature by combining word feature, global sentence feature, and learned query features. To analyze this, we conduct an ablation in Tab.~\ref{tab:text-ablation}. As in Tab.~\ref{tab:text-ablation}, when using only the word feature alone, the $\mathcal{J}\&\mathcal{F}$ score is 35.0\% (\ding{182}). Adding the sentence feature improves $\mathcal{J}\&\mathcal{F}$ from 35.0\% to 36.1\% (\ding{183}), while introducing the learned query feature improves it from 35.0\% to 36.6\% (\ding{184}), both enhancing the performance. When combining all features for expression representation, we achieve the best $\mathcal{J}\&\mathcal{F}$ score of 37.4\% (\ding{185}), validating the efficacy of our multi-cue representation for RSVIS.

\textbf{Ablation on context fusion in CGS.} We aggregate context information in CGS using normalized IoU scores to obtain the fused memory. We compare three fusion strategies, including simple addition with equal context weights, gated fusion with learned context weights, and our IoU-weighted fusion in Tab.~\ref{tab:fusion-ablation}. As shown, our strategy achieves the best $\mathcal{J}\&\mathcal{F}$ score of 37.4\% (\ding{184}). This result shows that the mask-quality scores provide a reliable signal for context aggregation.

\setlength{\columnsep}{6pt}%
\setlength\intextsep{-0pt}
\begin{wraptable}{r}{0.47\textwidth}
\setlength{\tabcolsep}{5.5pt}
	\centering
	\renewcommand{\arraystretch}{1.05}
    \caption{Ablation on $N_q$.}\vspace{-2mm}
	\resizebox{0.43\textwidth}{!}{
    \begin{tabular}{rcccc}
            \specialrule{1.5pt}{0pt}{0pt}
            & Num. of queries $N_q$ & $\mathcal{J}$ & $\mathcal{F}$ & $\mathcal{J}\&\mathcal{F}$ \\
            \hline\hline
            \ding{182} & $N_q=1$ & 34.9 & 38.1 & 36.5 \\
            \rowcolor{cyan!10}
            \ding{183} & $N_q=3$ & 36.1 & 38.7 & 37.4 \\
            \ding{184} & $N_q=5$ & 35.5 & 38.3 & 36.9 \\
            \ding{185} & $N_q=7$ & 35.3 & 38.1 & 36.7 \\
            \specialrule{1.5pt}{0pt}{0pt}
        \end{tabular}}
    \label{tab:query-ablation}
 \vspace{2mm}
\end{wraptable}
\textbf{Ablation on $N_q$ in the multi-cue representation.} In the multi-cue representation, we introduce $N_q$ learnable queries to capture fine-grained complementary cues in the expression. To study the impact of $N_q$, we conduct an ablation in Tab.~\ref{tab:query-ablation}. As shown in Tab.~\ref{tab:query-ablation}, using 3 learnable queries achieves the best $\mathcal{J}\&\mathcal{F}$ score of 37.4\% (\ding{183}), while using fewer (\ding{182}) or more (\ding{184} and \ding{185}) queries degrades performance, likely due to insufficient semantic coverage with smaller $N_q$ or redundant and noisy cues with larger $N_q$, respectively. Thus, we set $N_q$ to 3.

Due to limited space, additional details, results, and analyses are provided in the \textbf{\emph{appendix}}.

\section{Conclusion}

In this work, we present LD-RSVIS, a large-scale and diverse RSVIS benchmark. LD-RSVIS provides 3,536 videos with 1.09 million frames and diverse referring settings. To our best knowledge, LD-RSVIS is the largest benchmark for RSVIS and the first to support diverse referring settings. Moreover, to encourage future research, we introduce Cascade-RSVIS, a simple yet effective method for RSVIS. Our results demonstrate the advantages of our Cascade-RSVIS over existing methods.

\subsection*{AI use statement}

In this work, the generative AI tools are used \emph{solely} for language polishing and readability improvement. We do \emph{not} use the generative AI to generate datasets, produce experimental results, formulate mathematical claims, design experiments, implement methods, analyze results, or conduct the final interpretation of the findings. All AI-assisted text has been carefully reviewed, revised, and verified by the authors, who take full responsibility for the final content of this work.

\subsection*{Ethics Statement}

The construction of LD-RSVIS strictly follows ethical standards. LD-RSVIS is constructed from 15 publicly available surgical video sources, and can be used for research purpose only. All the sequences in LD-RSVIS remain subject to licenses, data-use agreements, and attribution requirements of their respective sources. Nevertheless, we understand that the availability or license of certain source videos might change in the future. Once any notification regarding this is received, we will take appropriate actions to handle it. 

\bibliography{iclr2027_conference}
\bibliographystyle{iclr2027_conference}

\clearpage
\appendix

\section*{Appendix}

\noindent
To better understand our LD-RSVIS and Cascade-RSVIS in this work, we provide additional details, analysis, results, and discussions as follows:

\vspace{-2mm}
\begin{itemize}
    \setlength{\itemsep}{3pt}
    \setlength{\parsep}{1pt}
    \setlength{\parskip}{1pt}
    \item \textbf{A \; \emph{Details of Surgical Instrument Categories and Procedures}} \\
    We provide the details of surgical instrument classes and procedures in LD-RSVIS.
    
    \item \textbf{B \; \emph{Construction Pipeline of LD-RSVIS and Additional Annotation Examples}} \\
    In this section, we present the detailed construction pipeline of LD-RSVIS and demonstrate more annotation examples.
    
    \item \textbf{C \; \emph{More Statistics and Annotation Quality Analysis}} \\
    We show additional statistics and annotation quality analysis of LD-RSVIS in this section.
    
    \item \textbf{D \; \emph{Details of Loss Function for Optimization}} \\
    We provide details of the loss function for optimization of Cascade-RSVIS.

    \item \textbf{E \; \emph{Model Complexity of Cascade-RSVIS}} \\
    We provide complexity analysis of Cascade-RSVIS in this section.
    
    \item \textbf{F \; \emph{Additional Results and Analysis}} \\
    We provide additional results and analysis in this section.
    
    \item \textbf{G \; \emph{Limitation of Cascade-RSVIS}} \\
    This section discusses the limitation of Cascade-RSVIS.
    
    \item \textbf{H \; \emph{Dataset Specification}} \\
    We discuss dataset specifications of LD-RSVIS.
\end{itemize}

\section{Details of Surgical Instrument Categories and Procedures}
\label{sec:supp-procedure-instrument}

LD-RSVIS aims at facilitating the development of general RSVIS. To this end, LD-RSVIS provides diverse surgical instrument classes and covers rich surgical procedures as described below. 

\subsection{Surgical Instrument Categories}

Specifically, LD-RSVIS contains 30 instrument classes, including ``\emph{Bipolar Forceps}'', ``\emph{Prograsp Forceps}'', ``\emph{Large Needle Driver}'', ``\emph{Monopolar Curved Scissors}'', ``\emph{Vessel Sealer}'', ``\emph{Grasping Retractor}'', ``\emph{Ultrasound Probe}'', ``\emph{Suction Instrument}'', ``\emph{Clip Applier}'', ``\emph{Grasper}'', ``\emph{Electric Hook}'', ``\emph{Irrigator}'', ``\emph{Scissors}'', ``\emph{Snare}'', ``\emph{Palpation Probe}'', ``\emph{Trocar}'', ``\emph{HF Coagulation Probe}'', ``\emph{Needle Probe}'', ``\emph{Drainage Instrument}'', ``\emph{Argon Beamer}'', ``\emph{Hook Clamp}'', ``\emph{Overholt Clamp}'', ``\emph{Sponge Clamp}'', ``\emph{LigaSure}'', ``\emph{Dissecting Forceps}'', ``\emph{Grasping Forceps}'', ``\emph{Suturing Needle}'', ``\emph{Clip}'', ``\emph{Catheter}'', and ``\emph{Needle Holder}''.

\subsection{Surgical Procedures}

LD-RSVIS covers 25 procedure categories, including ``\emph{Appendectomy}'', ``\emph{Cardiomyotomy}'', ``\emph{Cholecystectomy}'', ``\emph{Colectomy}'', ``\emph{Colon Cancer Resection}'', ``\emph{Colorectal Resection}'', ``\emph{Esophagectomy}'', ``\emph{Fundoplication}'', ``\emph{Gastrectomy}'', ``\emph{Gastrojejunostomy}'', ``\emph{Heller Myotomy}'', ``\emph{Hemicolectomy}'', ``\emph{Hernia Repair}'', ``\emph{Ladd's Procedure}'', ``\emph{Liver Resection}'', ``\emph{Rectal Resection}'', ``\emph{Rectal Prolapse Repair}'', ``\emph{Rectopexy}'', ``\emph{Sigmoidectomy}'', ``\emph{Splenectomy}'', ``\emph{Hysterectomy}'', ``\emph{Nephrectomy}'', ``\emph{Roux-en-Y Gastric Bypass}'', ``\emph{Radical Prostatectomy}'', and ``\emph{Proctocolectomy}''.

\section{Construction Pipeline of LD-RSVIS and Additional Annotation Examples}
\label{sec:supp-construction}

\begin{figure*}[!t]
    \centering
    \includegraphics[width=0.95\textwidth]{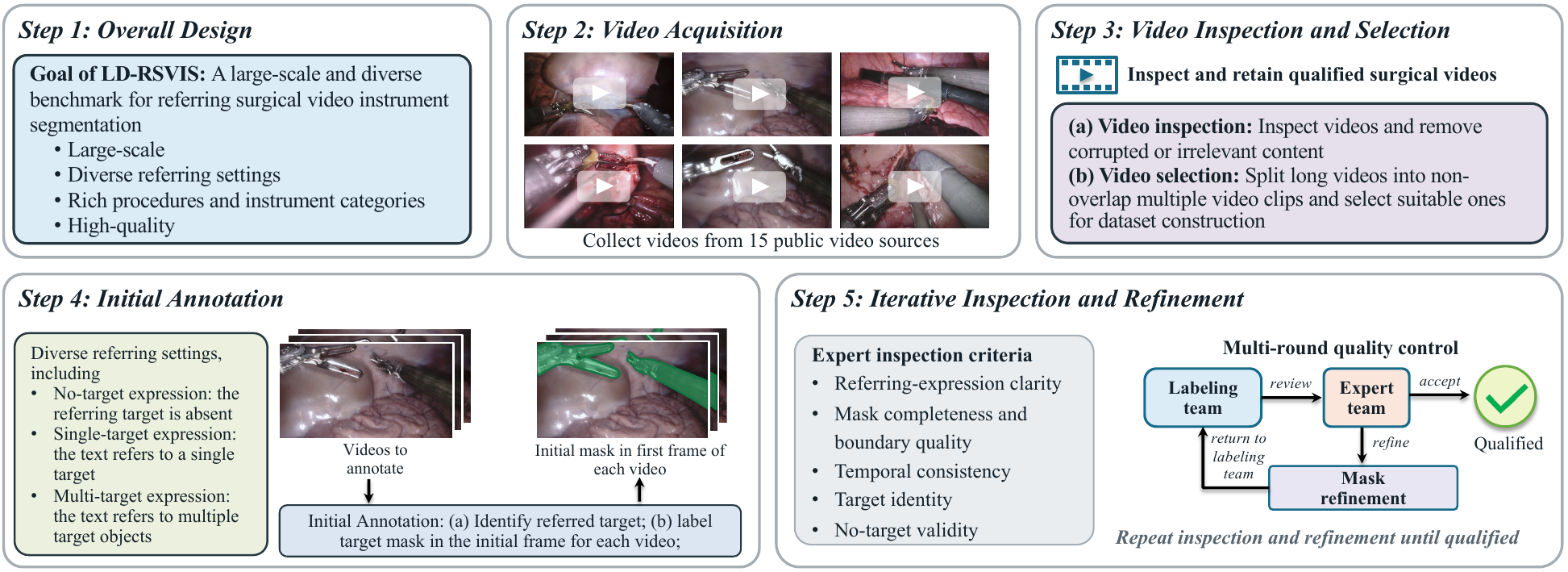}
    \caption{Illustration of the construction pipeline of LD-RSVIS.}
    \label{fig:supp-construction-pipeline}
\end{figure*}

\begin{figure}[!t]
    \centering
    \includegraphics[width=0.95\textwidth]{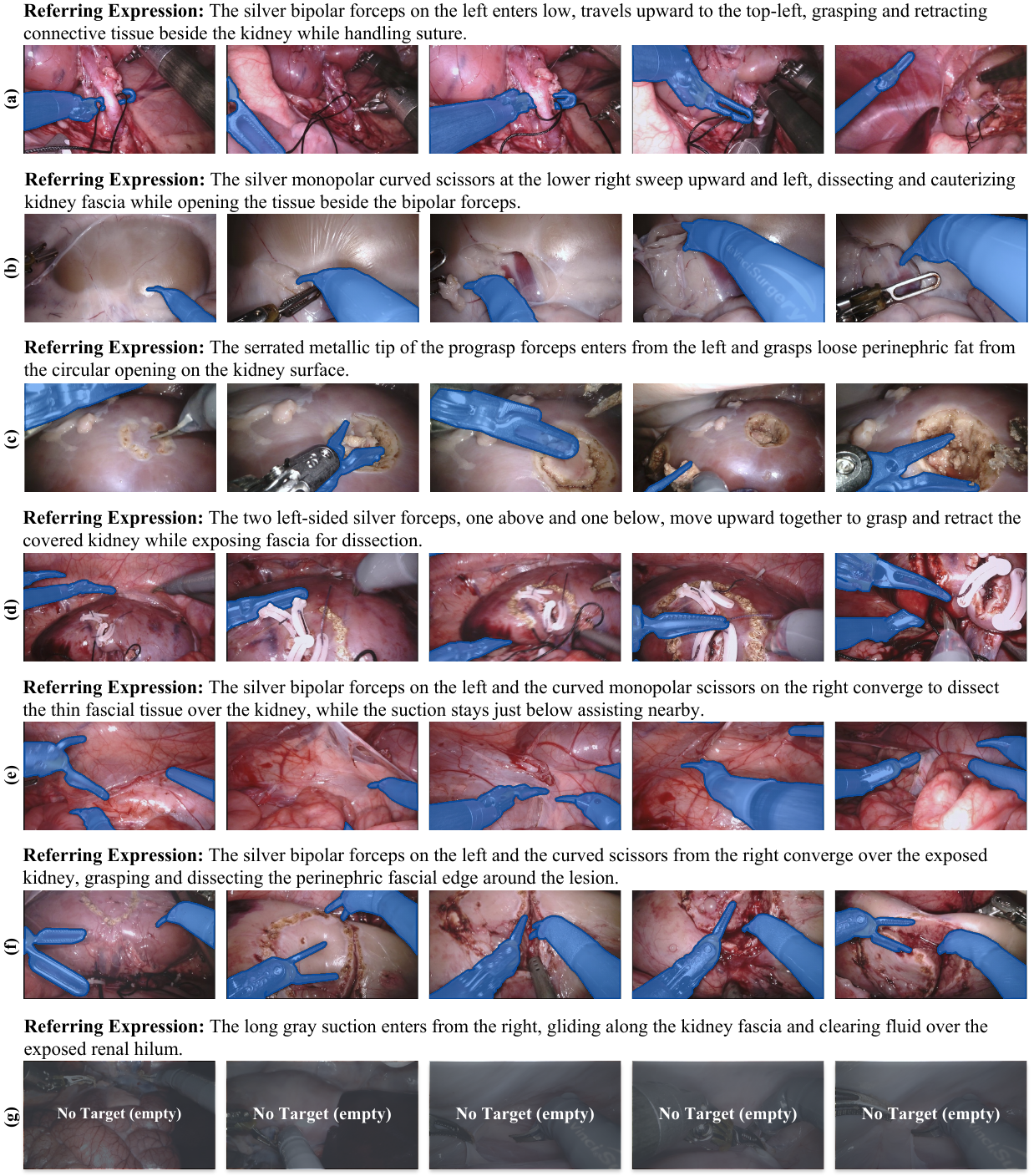}
    \caption{Additional annotation examples in LD-RSVIS.}
    \label{fig:supp-annotation-examples}
\end{figure}

The construction of LD-RSVIS contains five steps: \textbf{(i) \emph{Overall design}.} We first determine the scope of referring surgical video instrument segmentation and the criteria for selecting source datasets, procedures, instrument classes, and referring settings. \textbf{(ii) \emph{Video acquisition}.} We then collect videos from 15 public source datasets, covering surgical scene understanding, phase and action recognition, workflow analysis, general surgical visual understanding, and surgical video pre-training. \textbf{(iii) \emph{Video selection}.} We carefully inspect videos from different sources and remove corrupted or irrelevant content. Qualified long videos are divided into representative continuous sequences that preserve diverse instrument appearances, surgical stages, and temporal conditions while avoiding excessive temporal redundancy. \textbf{(iv) \emph{Initial annotation}.} For each retained sequence, experts generate referring expressions and offer initial frame-level masks. \textbf{(v) \emph{Iterative inspection and refinement}.} The initial masks are extended to consistent spatio-temporal masklets by the labeling team and then reviewed by experts for mask completeness, boundary quality, temporal consistency, and target identity. Annotations that do not satisfy the quality requirements are returned to the labeling team for refinement. This process is repeated until all annotations are qualified. The construction pipeline is illustrated in Fig.~\ref{fig:supp-construction-pipeline}. In Fig.~\ref{fig:supp-annotation-examples}, we demonstrate more annotation samples of LD-RSVIS.

\section{More Statistics and Annotation Quality Analysis}
\label{sec:supp-statistics}

\begin{table}[!t]
    \centering
    \setlength{\tabcolsep}{4pt}
    \renewcommand{\arraystretch}{1.02}
    \caption{Composition of selected data from source datasets for constructing LD-RSVIS.}\vspace{-2mm}
    \resizebox{\textwidth}{!}{%
    \begin{tabular}{rcccc}
        \Xhline{1.2pt}
        \textbf{Source Dataset} & \textbf{Task of the Source Dataset} & \textbf{Selected Videos} & \textbf{Selected Frames} \\
        \hline\hline
        EndoVis17~\citep{Allan2019EndoVis17} & surgical scene segmentation and/or detection & 18 & 3,000 \\
        EndoVis18~\citep{Allan2020EndoVis18} & surgical scene segmentation and/or detection & 15 & 2,235 \\
        Cholec80~\citep{twinanda2016endonet} & surgical workflow understanding & 318 & 98,620 \\
        Endoscapes2023~\citep{murali2023endoscapes} & surgical scene segmentation and/or detection & 181 & 55,840 \\
        HeiChole~\citep{wagner2023comparative} & surgical workflow understanding & 170 & 52,460 \\
        PhaKIR~\citep{rueckert2026video} & surgical phase and action recognition  & 60 & 18,640 \\
        AutoLaparo~\citep{wang2022autolaparo} & surgical scene segmentation and/or detection & 218 & 67,420 \\
        MultiBypass140~\citep{lavanchy2024challenges} & surgical phase and action recognition & 654 & 202,350 \\
        HeiCo~\citep{maier2021heidelberg} & surgical scene segmentation and/or detection & 442 & 136,840 \\
        ESAD~\citep{bawa2021saras} & surgical phase and action recognition & 169 & 52,180 \\
        GraSP~\citep{ayobi2025pixel} & surgical scene segmentation and/or detection & 238 & 73,420 \\
        SAR-RARP50~\citep{psychogyios2023sar} & surgical phase and action recognition & 49 & 14,210 \\
        SurgVU~\citep{zia2026surgical} & surgical task and activity recognition & 393 & 121,360 \\
        hSDB-instrument~\citep{yoon2021hsdb} & surgical scene segmentation and/or detection & 115 & 35,480 \\
        SurgeYoutube~\citep{JASPERS2026103873} & large-scale surgical video pre-training & 496 & 153,307 \\
        \hline
        \textbf{Total} & -- & \textbf{3,536} & \textbf{1,09 Million} \\
        \Xhline{1.2pt}
    \end{tabular}}
    \label{tab:supp-source-statistics}
\end{table}

\textbf{Selected data from source datasets.} Our LD-RSVIS comprises 3,536 surgical videos and 1.09 million frames selected from 15 publicly available source datasets. Tab.~\ref{tab:supp-source-statistics} summarizes the selected data from each source dataset and their contributions to LD-RSVIS. The source pool includes both standard surgical video benchmarks and datasets developed for broader surgical video understanding, providing diverse coverage of surgical procedures, instrument categories, and temporal contexts.

\setlength{\columnsep}{6pt}%
\setlength\intextsep{-0pt}
\begin{wraptable}{r}{0.5\textwidth}
\setlength{\tabcolsep}{5pt}
	\centering
	\renewcommand{\arraystretch}{1.05}
    \caption{Distribution of different types of referring expression in LD-RSVIS.}\vspace{-2mm}
	\resizebox{0.45\textwidth}{!}{
    \begin{tabular}{rcccc}
            \specialrule{1.5pt}{0pt}{0pt}
             & \textbf{MTE} & \textbf{STE} & \textbf{NTE} & \textbf{Total} \\
            \hline\hline
            LD-RSVIS$_\text{Tra}$  & 11,412 & 28,171 & 7,917 & 47,500 \\
            LD-RSVIS$_\text{Tst}$      & 2,732  & 6,573 & 1,861 & 11,166 \\
            LD-RSVIS  & 14,144 & 34,744 & 9,778 & 58,666 \\
            \specialrule{1.5pt}{0pt}{0pt}
        \end{tabular}}
    \label{tab:exp}
 \vspace{2mm}
\end{wraptable}
\textbf{Distribution of different types of expressions.} Tab.~\ref{tab:exp} summarizs the expression composition of LD-RSVIS. Specifically, the training set LD-RSVIS$_\text{Tra}$ contains 11,412 multi-target expressions (MTE), 28,171 single-target expressions (STE), and 7,917 no-target expressions (NTE). The test set LD-RSVIS$_\text{Tst}$ contains 2,732 MTE, 6,573 STE, and 1,861 NTE. Together, our LD-RSVIS comprises 58,666 referring expressions, with 34,744, 14,144, and 9,778 STE, MTE, and NTE, respectively.

\textbf{Annotation quality analysis.} To analyze our annotation quality, we randomly choose 50 sequences from our LD-RSVIS and ask an independent group of external experts to inspect and re-label them. Then, we compute the Intersection over Union (IoU) of new and original masklets. The IoU of these selected sequences is 0.93, validating the quality and reliability of our annotations.

\section{Details of Loss Function for Optimization}
\label{sec:supp-training}

Cascade-RSVIS contains two stages of TCM and CGS. Given a video sequence $\mathcal{I}=\{I_i\}_{i=1}^{L}$ with $L$ frames and a referring expression, both stages predict masks. During training, given the groundtruth, we follow SAM 2~\citep{ravi2025sam} and utilize multiple losses, including focal loss, dice loss, IoU loss, and binary cross entropy loss to compute the frame-level loss $\mathcal{L}_{\text{TCM}}^{i}$ and $\mathcal{L}_{\text{CGS}}^{i}$ for TCM and CGS, respectively (please refer to these losses in~\citep{ravi2025sam} for details). The losses over the entire video sequence are computed as $\mathcal{L}_{\text{TCM}}=\sum_{i=1}^{L}\mathcal{L}_{\text{TCM}}^{i}$ and $\mathcal{L}_{\text{CGS}}=\sum_{i=1}^{L}\mathcal{L}_{\text{CGS}}^{i}$. In this work, we adopt a stage-wise optimization strategy. Specifically, we first optimize TCM using $\mathcal{L}_{\text{TCM}}$. After this, CGS is optimized using $\mathcal{L}_{\text{CGS}}$ with the same input together with outputs from TCM.

\section{Model Complexity of Cascade-RSVIS}

We train Cascade-RSVIS using two NVIDIA A100 (40G) GPUs for around 20 hours. The resolution of video frames for both training and inference is $896\times896$. Cascade-RSVIS contains 292M total parameters. During inference, the video is divided into eight-frame clips and processed on a single NVIDIA A100 GPU. The peak GPU memory requirement for running Cascade-RSVIS is 16~GB, and its inference speed is 25 frames per second (\emph{fps}).

\section{Additional Results and Analysis}
\label{sec:supp-additional-results}

\begin{table}[!t]
    \centering
    \setlength{\tabcolsep}{5pt}
    \renewcommand{\arraystretch}{1.0}
    \caption{Comparison on existing public RSVIS benchmarks. oIoU and mIoU denote Overall IoU and Mean IoU, respectively.}\vspace{-2mm}
    \resizebox{0.98\textwidth}{!}{%
    \begin{tabular}{rcccccccccccc}
        \Xhline{1.2pt}
        & \multicolumn{6}{c}{\textbf{EndoVis-RS}}
        & \multicolumn{6}{c}{\textbf{Ref-EndoVis (tool)}} \\
        \cmidrule(lr){2-7} \cmidrule(lr){8-13}
        & \multicolumn{3}{c}{RS17} & \multicolumn{3}{c}{RS18}
        & \multicolumn{3}{c}{Ref17} & \multicolumn{3}{c}{Ref18} \\
        \cmidrule(lr){2-4} \cmidrule(lr){5-7} \cmidrule(lr){8-10} \cmidrule(lr){11-13}
        \textbf{Methods}
        & oIoU & mIoU & mAP
        & oIoU & mIoU & mAP
        & $\mathcal{J}$ & $\mathcal{F}$ & $\mathcal{J}\&\mathcal{F}$
        & $\mathcal{J}$ & $\mathcal{F}$ & $\mathcal{J}\&\mathcal{F}$ \\
        \hline\hline
        VIS-Net~\citep{wang2024video}\textsubscript{[T-MI'24]} & 65.5 & 70.1 & 53.8 & 74.2 & 72.3 & 60.1 & 61.4 & 61.1 & 61.2 & 68.6 & 68.2 & 68.4 \\
        SurgRef~\citep{wei2026moves}\textsubscript{[AAAI'26]} & 66.4 & 70.8 & 54.7 & 74.8 & 73.4 & 60.8 & 70.8 & 69.6 & 70.2 & 76.1 & 74.7 & 75.4 \\
        ReSurgSAM2~\citep{liu2025resurgsam2}\textsubscript{[MICCAI'25]} & 68.7 & 72.9 & 57.2 & 76.5 & 75.3 & 62.9 & 77.8 & 77.7 & 77.7 & 80.9 & 80.3 & 80.6 \\
        ReferMo~\citep{liang2026long}\textsubscript{[CVPR'26]} & 69.4 & 73.4 & 57.6 & 77.2 & 75.8 & 63.3 & 78.2 & 78.8 & 78.5 & 81.0 & 81.4 & 81.2 \\
        FlowRVS~\citep{wang2026deforming}\textsubscript{[ICLR'26]} & 69.0 & 74.0 & 58.1 & 76.8 & 76.4 & 64.0 & 77.8 & 80.6 & 79.2 & 80.8 & 82.8 & 81.8 \\
        \hline
        \rowcolor[HTML]{e9f7ef}
        Cascade-RSVIS (ours) & \textbf{71.6} & \textbf{75.9} & \textbf{59.8} & \textbf{79.0} & \textbf{78.1} & \textbf{65.3} & \textbf{80.2} & \textbf{81.0} & \textbf{80.6} & \textbf{82.1} & \textbf{83.1} & \textbf{82.6} \\
        \Xhline{1.2pt}
    \end{tabular}}
    \label{tab:supp-surgical-benchmarks}
\end{table}

\subsection{Comparison on Existing RSVIS Benchmarks}

To evaluate the generalization ability of Cascade-RSVIS on existing public surgical RSVIS benchmarks, we conduct experiments on EndoVis-RS~\citep{wang2024video} and Ref-EndoVis~\citep{liu2025resurgsam2}, which are the two publicly available benchmarks with released data. We compare Cascade-RSVIS with three surgical RSVIS methods, including VIS-Net~\citep{wang2024video}, SurgRef~\citep{wei2026moves}, and ReSurgSAM2~\citep{liu2025resurgsam2}, as well as two recent general RVOS methods, ReferMo~\citep{liang2026long} and FlowRVS~\citep{wang2026deforming}. For Ref-EndoVis, we report results on tool expressions only, since LD-RSVIS focuses on surgical instruments and their parts rather than tissues. We follow the original evaluation protocols of each benchmark, adopting Overall IoU (oIoU), Mean IoU (mIoU), and mAP for EndoVis-RS, and $\mathcal{J}$, $\mathcal{F}$, and $\mathcal{J}$\&$\mathcal{F}$ for Ref-EndoVis.

Tab.~\ref{tab:supp-surgical-benchmarks} reports the comparison results. ReferMo and FlowRVS transfer favorably to both surgical benchmarks and outperform the surgical RSVIS methods in terms of mAP and $\mathcal{J}\&\mathcal{F}$ across all four subsets, with FlowRVS achieving the strongest aggregate results among the compared approaches. Their margins over ReSurgSAM2 are comparatively smaller on these benchmarks than under the STE setting of LD-RSVIS. This is because the existing benchmarks focus on whole-instrument targets in a limited number of surgical scenes, which closely matches the target-initialization and long-term tracking design of ReSurgSAM2, whereas LD-RSVIS additionally includes fine-grained part-level expressions and substantially greater scene diversity. Nevertheless, Cascade-RSVIS obtains mAP scores of 59.8\% and 65.3\% on RS17 and RS18 of EndoVis-RS, exceeding FlowRVS by 1.7\% and 1.3\%, respectively. On Ref-17 and Ref-18 of Ref-EndoVis (tool), it achieves $\mathcal{J}\&\mathcal{F}$ scores of 80.6\% and 82.6\%, improving on FlowRVS by 1.4\% and 0.8\%. These results consistently demonstrate the effectiveness of Cascade-RSVIS for referring surgical video instrument segmentation.

\subsection{Qualitative Comparison}

\begin{figure}[!t]
    \centering
    \includegraphics[width=\textwidth]{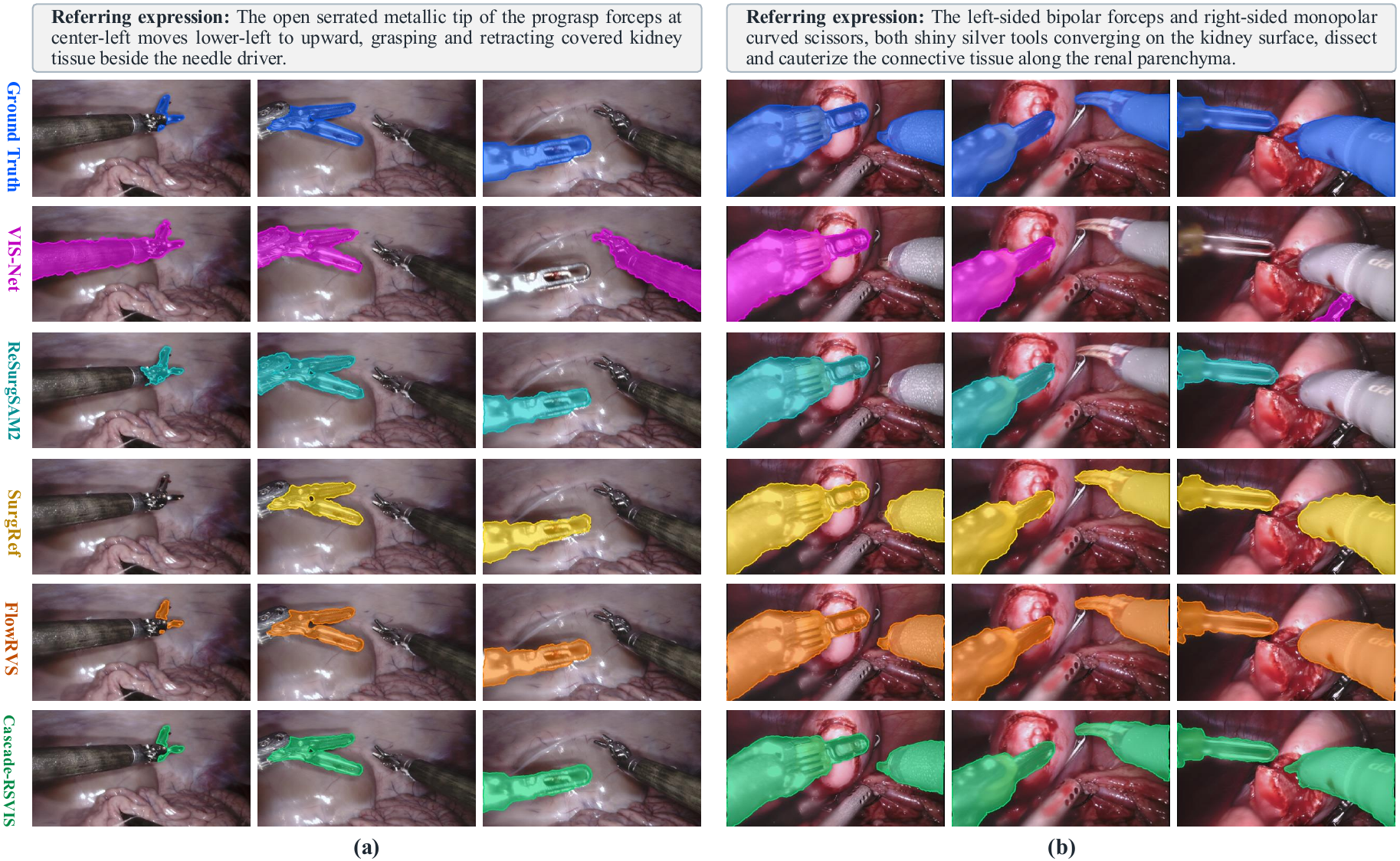}
    \caption{Qualitative comparison of Cascade-RSVIS with representative  methods on LD-RSVIS. Image (a)  shows the results of a single target case, while image (b) shows a multi-target case.}
    \label{fig:supp-qualitative-comparison}
\end{figure}

To provide further qualitative analysis of Cascade-RSVIS, we compare its segmentation results with four representative RSVIS methods in Fig.~\ref{fig:supp-qualitative-comparison}. Fig.~\ref{fig:supp-qualitative-comparison}-left shows a challenging single-target example, where the referred instrument tip occupies only a small region and moves substantially across the video. VIS-Net tends to segment the complete parent instrument and switches to a neighboring instrument in the late frame, ReSurgSAM2 exhibits leakage into the adjacent instrument part, and SurgRef misses the target in the early frame. FlowRVS localizes the referred part more accurately but still exhibits local boundary deviations. Figure~\ref{fig:supp-qualitative-comparison}-right further shows a multi-target example in which two instruments are referred to among four simultaneously visible instruments. The compared methods either omit one referred target, switch to an unreferred distractor, or lose fine instrument regions across frames; FlowRVS retains both target identities but loses part of the fine regions in the late frame. In contrast, Cascade-RSVIS consistently identifies the referred target set and produces accurate masks throughout both videos, demonstrating the effectiveness of target-context mining and context-guided spatial-temporal segmentation.

\section{Limitation of Cascade-RSVIS}
\label{sec:supp-limitations}

In this work, we propose a simple yet effective Cascade-RSVIS to encourage more future research on RSVIS. Despite promising results, Cascade-RSVIS has two limitations. First, it may degrade in highly challenging surgical scenes. In these scenes, very small instrument parts may be heavily occluded, and multiple instruments with similar appearances may occur simultaneously, resulting in lower segmentation performance. To handle these cases, special strategies such as leveraging richer spatial context through explicit instrument-instrument interaction or target trajectory cues are needed. Second, the current framework does not explicitly model instrument motion. Although temporal information is implicitly captured through video features, explicit motion cues, such as motion-aware correspondence or instrument trajectories, may provide complementary information for distinguishing visually similar instruments and maintaining target consistency under rapid movement or occlusion. Exploring such motion-aware modeling is an interesting direction for future work. Considering that our current goal is to provide a feasible baseline for RSVIS, we leave the above explorations to future work.

\section{Dataset Specification}
\label{sec:supp-ethics}

\textbf{Annotator Protection.} To safeguard annotator well-being, all participants provide informed consent and are notified in advance that the surgical videos may contain potentially distressing content. Annotators are allowed to skip any video that causes discomfort and may withdraw from the annotation process at any time.

\textbf{Maintenance.} LD-RSVIS will be hosted on GitHub and/or Hugging Face. This allows us to conveniently check feedback from the community and improve LD-RSVIS through necessary maintenance and updates. In addition, all the annotations and results in this work will be made publicly available. Our ultimate goal is to offer a long-term and stable platform for the community to foster research on referring surgical video instrument segmentation.

\textbf{Responsible Use of LD-RSVIS.} LD-RSVIS is intended to support research on referring surgical video instrument segmentation and is released for research purposes only. Users should be aware that, due to potential biases inherited from the source datasets and the annotation process, the dataset may exhibit imbalances across surgical procedures, institutions, acquisition devices, and instrument categories. It is worth noting that, the clinical validation of this work is required before their deployment in real-world surgical settings.

\textbf{Potential Limitation.} A potential 
limitation of LD-RSVIS is the substantial cost and complexity of annotation. Accurately segmenting referred surgical instruments across video sequences, in diverse multi-target, single-target and no-target settings, requires detailed spatiotemporal annotations. This process is labor-intensive and time-consuming, making it difficult to scale the benchmark rapidly to more videos, surgical procedures, and clinical domains.

\end{document}

%% file: math_commands.tex
\usepackage{amsmath,amsfonts,bm}

\def\eqref#1{equation~\ref{#1}}

\def\1{\bm{1}}

\DeclareMathAlphabet{\mathsfit}{\encodingdefault}{\sfdefault}{m}{sl}
\SetMathAlphabet{\mathsfit}{bold}{\encodingdefault}{\sfdefault}{bx}{n}

